\documentclass[a4paper, 11pt]{amsart}
\usepackage{amsfonts, amsthm, amssymb, amsmath}
\usepackage{mathrsfs,array}
\usepackage{hyperref}
\usepackage{verbatim}
\usepackage[latin1]{inputenc}

\usepackage[titletoc]{appendix}
\usepackage{color}

\theoremstyle{definition}

\title{D\MakeLowercase{eep}A\MakeLowercase{lgebra - an outline of a program}}
\date {\today}
\author{Przemys\l aw Chojecki}

\address{Przemys\l aw Chojecki\\
University of Oxford\\ 
and Polish Academy of Sciences
}
\email{prz.chojecki@gmail.com}
\urladdr{http://people.maths.ox.ac.uk/chojecki/}

\thanks{These ideas were conceived while the author was a research fellow at the University of Oxford. We thank Wojciech Zaremba (OpenAI) for pointing us to \cite{deep} which triggered writing this article and Maciej Zdanowicz (University of Warsaw) for numerous discussions which helped the author clarify his ideas.}

\begin{document}

\begin{abstract}

We outline a program in the area of formalization of mathematics to automate theorem proving in algebra and algebraic geometry. We propose a construction of a dictionary between automated theorem provers and (La)TeX exploiting syntactic parsers. We describe its application to a repository of human-written facts and definitions in algebraic geometry (The Stacks Project). We use deep learning techniques.
\end{abstract}

\maketitle

\section{Introduction}

\noindent Mathematics is a basis for the modern world. Not just simple mathematics, but also complex proofs underpin current breakthroughs in technology and science. Much of the physics depends on mathematical proofs. Proofs in mathematics itself are ubiquitous and many new mathematical theorems are proved daily. This leads to a specialization and a loss of a global picture for most of mathematicians. The formalization of mathematics started taking place to circumvent this difficulty and make sure that modern mathematics stands on firm grounds (cf. \cite{hales}).

\medskip 

\noindent The largest single piece of formalized mathematics to this day is a proof of the Feit-Thompson Odd Order Theorem, done by collaborative efforts of 15 mathematicians over a period of six years (\cite{odd}). The proof itself which spans 250 pages of mathematics was formalized into more than 150,000 lines of code with roughly 4,000 definitions and 13,000 theorems. The research group developed along the way many reusable libraries in the COQ proof assistant (\cite{coq}).

\medskip

\noindent It is obvious that if the project of formalizing mathematics is to be completed, it has to proceed much faster and use more effectively human power - especially if we want to use machines for proving new theorems. In \cite{deep} authors name two main bottlenecks: 

\medskip

\noindent (1) lack of automated methods for formalization (\textit{autoformalization}); \newline
(2) lack of strong automated methods to fill in the gaps in already formalized human-written proofs. 

\medskip

\noindent A basis for the research in \cite{deep}, where the authors deal with the second bottleneck, is Mizar Mathematical Library (MML) which contains over 50,000 lemmas. The authors have used deep learning methods to select premises for the automatic theorem proving and automatically prove as many theorems as possible using at each step all previous theorems and proofs. To some extent this solves (2), though much more optimalization work is needed to attain level of human effectiveness.  

\medskip

\noindent In this work, we focus on the first bottleneck. We propose a program to automate a formalization of large parts of modern algebraic geometry using deep learning techniques run on well-chosen repositories of human-written mathematical facts (The Stacks Project \cite{stacks}). The main problem is the construction of a dictionary between human-written proofs in (La)TeX and Interactive Theorem Provers. Here we outline our program and lay a theoretical basis for it. We report on our progress on implementing it in subsequent papers and in \cite{deepalgebra}. Our main theoretical input is to use syntactic parsers run on The Stacks Project to correctly translate \cite{stacks} into COQ, and then use hammers to verify its validity in COQ/E, and possibly reprove some of it by automated methods. Eventually this approach should lead to proving new theorems in algebraic geometry purely by machines.

\medskip

\noindent As the last remark in the introduction, we notice that the formalization of mathematics and automatic theorem proving is important, because it can be viewed as a toy model for a harder problem, namely constructing an AI with an ability to write a self-correcting code (this is listed as one of the special projects in OpenAI project, see \cite{openai}). We believe that bringing theorem proving by AI to the human-level is a necessary step (and a very important one) towards tackling this harder problem.

\section{The Stacks Project}

\noindent Algebraic geometry is one of the pillars of modern mathematics. Its main object of study is algebraic varieties (vanishing loci of sets of polynomials) and maps between them. It builds upon standard algebra and study of rings and modules, and as such is highly symbolic. Because of that, we believe it is easier to formalize it rather than more analytic parts of mathematics where reasoning is more ad hoc and proofs can use tricky approximations, which seem hard to teach to a computer. On the other hand, the problem with algebraic geometry is the level of abstraction and the amount of terms which one needs to formulate the problems correctly. 

\medskip 

\noindent When trying to formalize human-written proofs with a goal of training neural networks on them, one has to be sure that proofs are correct in the first place. In other words, one has to choose sources well. Mathematicians often write proofs in informal ways, leaving gaps to readers which are assumed to be experts in a given field as well.

\medskip

\noindent That is why we propose as our training source the Stacks Project (\cite{stacks}). This is a repository of facts and theorems in algebra and algebraic geometry, which starts from the basic material and goes up to the current developments in the field. It is still actively developped with the help of dozens of volunteers and currently contains 509,794 lines of code and 5,347 pages (as we write).

\medskip

\noindent Its huge advantage is that it is completely self-contained. There are no references to outside sources. Every fact is proved and can be traced to the axioms introduced at the beginning. The only problem for our goal is that it is written by humans and for humans.

\section{Dictionary}

\noindent In order to formalize this amount of mathematics (and go beyond it) one needs to automate the process. We remind a reader that our goal is to develop an AI which could prove new theorems in algebraic geometry by itself. To do that, one firstly needs to translate the source (The Stacks project in our case) to one of \textbf{Interactive Theorem Provers} (ITPs) such as COQ \cite{coq} or Mizar \cite{mizar} \cite{mizar2} \footnote{Other standard ITPs include HOL (Light), Isabelle and ACL2.}, and then use an \textbf{Automatic Theorem Prover} (ATP), for example E \cite{sch1} \cite{sch2} \footnote{Other standard ATPs include Vampyre, Z3 and JProver.}, together perhaps with some deep learning techniques to facilate proving more complex theorems.

\medskip

\noindent The first step of our DeepAlgebra program is to construct a \textbf{dictionary} between (La)TeX files and COQ/Mizar, which would translate .tex files into ITPs and vice versa. While one direction is relatively easy (ITP to .tex) as one can use fairly simple, hard-coded language, the other direction is much less trivial. This is due to the fact to which we have alluded before, that human-written mathematics happens in general to be informal with many gaps left to readers. Computer needs to be taught which sentences and words can be ignored as informal discussions and which are crucial to on-going proofs, as mathematical papers are not constructed in the form of "theorem, proof, theorem, proof", but often contains a valuable discussion outside of it. 

\medskip

\noindent The other problem is to correctly implement the abstraction in the form of Types (COQ). Algebra tends to use many words for the same thing and correctly storing it is crucial. For example computer needs to know that a "reduced scheme of finite type over a field" is the same thing as an "algebraic variety", while both characterizations give slightly different perspectives.  

\medskip

\noindent When trying to formalize The Stacks Project, most of these problems do not appear (or are easier) as the text itself is written in the form of definitions, followed by lemmas, followed by theorems. Thus translating .tex file into COQ in this case needs much less work than with the general mathematical paper \footnote{Partly because of the clear dependancy graph of definitions and theorems used in the Stacks Project, cf. section Hammers.}, and we shall report on our progress elsewhere. However this does not solve the general problem, which is needed in order to formalize current mathematical work. General mathematical papers tend to be worse structured with respect to machines.

\medskip

\noindent	Let us observe that one can divide dictionaries into two categories:

\medskip

\noindent (1) \textbf{automated}, where no human-help is needed to make a translation between (La)TeX and ITPs; 

\medskip

\noindent (2) \textbf{semi-automated}, which are assisted by a human, to correct mistakes and fill in the gaps in human-written proofs, which could not be filled by a computer.

\medskip

\noindent One of general goals in the field of \textbf{Automated Reasoning in Large Theories} (ARLT) \cite{arlt} should be creating a perfect automated dictionary, which translates .tex files to COQ/Mizar without information losses \footnote{Of course ARLT does not necessarily mean mathematics and thus in different scientific disciplines by a dictionary we mean a (semi)-automated way of passing between human-written science and according formal verificators.}. Nevertheless for our purposes of formalizing The Stacks Project semi-automated dictionary would be enough and is much easier to construct.  

\medskip

\noindent We remark that the problem of constructing a dictionary does not appear when one starts with already formalized proofs like Mizar Mathematical Library (as in  \cite{deep}). This however is very limited and in the end one needs to find ways to automatically formalize human-written proofs to keep up with the developments of modern mathematics and eventually surpass it using AI. Despite many efforts to optimize ITPs/ATPs and their theorem proving capabilities, there are currently no dictionaries existing.

\section{Building a dictionary}

\noindent Creating a dictionary in the above sense can be viewed as a \textbf{Natural Language Processing} (NLP) problem, where one tries to pass between human-written mathematics and ITPs. This approach offers plethora of methods to choose from. We present one which we plan to implement. 

\medskip

\noindent For us a \textbf{Type} is an abstract mathematical term (e.g. "group"), while a \textbf{variable} is an object of certain Type (e.g. in a sentence "Let $G$ be a group" $G$ is a variable of Type "group"). By \textbf{relation} we mean a first order logical sentence involving Types and variables. By a \textbf{library} we will mean an already constructed repository of defined objects and proven formalized theorems (in COQ), which can be used to prove new theorems. By an \textbf{environment} we mean a statement of a lemma or a theorem, or a proof (so any piece in (La)TeX in a form of $\backslash$ begin\{\} ... $\backslash$ end\{\}). This terminology is consistent with COQ.

\medskip

\noindent \textbf{Method:} We build a dictionary by using a syntactic parser to identify Types and variables together with relations between them, which we then transfer directly to COQ. Issues which usually arise in machine translations like idioms or complex grammar, rarely appear in mathematical texts, hence a syntactic parser together with hard-coded translation of certain phrases (and environments) is enough to build a dictionary.

\medskip

\noindent \textbf{Implementation:} When analysing a .tex file we differentiate between (mathematical) English text and formulas occuring most commonly between dollar signs. Processing formulas from (La)TeX to COQ is relatively easy - apart from complicated diagrams which have to be split into direct formulas \footnote{Instead of a diagram we will have a bunch of sentences of the form "$f: X \rightarrow Y$ is a morphism such that..."} - we can basically rewrite (La)TeX code into COQ one to one (i.e. a rule-based approach to machine translation). 

\medskip

\noindent The real problem is putting a text into COQ. A typical mathematical sentence will look like "$A$ has property $P$, because $B$". In order to put it into COQ, we have to identify Types and variables (what kind of objects we are considering; here $A$ can be either a Type or a variable), then identify logical dependencies (Type of $A$ determined by $B$ implies $P$). The actual verification of a sentence in COQ is the next step with which we deal in the next section. Building a dictionary amounts to translating .tex file into COQ before verifying it formally.

\medskip 

\noindent In a mathematical sentence we first identify objects between dollar signs and nouns as potential Types and variables. This can be done using a modified version of spacy.io \cite{sp} \footnote{To be precise - we need a syntactic parser for syntax analysis; recently Google went open-source with its syntactic parser SyntaxNet and implemented it into TensorFlow \cite{syntax}; spacy.io builds upon this Natural Language Understanding tool.} where one treats any \$...\$ expressions with no operators (=, $<$, $>$, etc.) as words (and not formulas). Spacy.io gives a clear syntax decomposition and a parse tree (a dependency graph of a sentence). We look at nouns and \$...\$ objects and analyse which objects are already defined (as Types or variables) and which objects are not; this is done by a direct check with the library of terms we are maintaining. We conclude that those which are not known to our COQ library are being defined right now. From the syntax analysis done by spacy.io we get what is defined by what - we obtain a dependency graph of nouns (objects) defined by other nouns (objects) together with accompanying adjectives (properties). Verbs seem to not play a role in COQ and are only needed to indicate a dependence relation ("$A$ has property $P$"), which we have already exploited through a syntactic parser.

\medskip

\noindent In a basic sentence like "Let $X$ be a reduced scheme.", spacy.io tells us that "reduced" is a property of a "scheme" and $X$ refers to "be" which points to "scheme". Thus we define $X$ as variable of Type "reduced scheme", which itself is a sub-Type of Type "scheme". 

\medskip

\noindent We remark that one has to distinguish between defining new variables (like $X$) and Types (like "reduced scheme"). A priori it seems natural to define any \$...\$ as a variable and any unidentified nouns as Types \footnote{One has to be careful with non-technical terms, but in the worst case they can also be defined as separate Types. The way out is to only start adding new Types/variables once DeepAlgebra finds a "Let ... be" expression or a similar one.}. This should give correct output most of the time. However for a more tricky example consider an adjective "\$p\$-adic", mostly written in this way; here \$p\$ does not define a new variable but is a part of an adjective. Fortunately in a sentence "Let \$X\$ be a \$p\$-adic scheme" the word "\$p\$-adic" is correctly seen by spacy.io as an adjective related to "scheme". Thus the solution seems to be looking only for words classified as nouns by spacy.io, when looking for variables and Types. 

\medskip

\noindent A different problem to deal with is to interpret side remarks in mathematical texts which give some new perspective on a problem but does not necessarily give any input into proofs. The way to overcome it is to ignore any sentences which do not contain "triggers" like "Let ... be", "because", "since", "thus", \$...\$, etc. or actually ignore everything outside of a rigid structure Lemma, Proof, Theorem, Proof - this can be done with the Stacks Project. Another difficulty is to transfer whole environments (Theorem, Lemma, Proof, etc.) from .tex to COQ, but this is relatively simple as COQ uses the same environments as ordinary mathematical texts thus this can be hard-coded.

\medskip

\noindent Once the translation into COQ is done we want to verify whether the sentence is valid; in our first example we want to conclude that $B$ implies $P$. This is done in COQ using E and we discuss it in the next section. This takes place purely in COQ as we have formalized everything at this stage.

\section{Hammers}

\noindent Creating a dictionary is only the first step in automatic theorem proving - it allows to amass training material for an AI (neural networks). The next step is to actually perform an automated reasoning. Recent years show some activity in this area. The main activity concerned recreating some of the already formalized mathematics by using certain premises (lemmas/theorems). In DeepAlgebra project we will apply these already established techniques to the Stacks Project, after we formalize it using our dictionary. We quickly survey most up-to-date techniques in automatic theorem proving.

\medskip

\noindent The main way to pass between ITPs and ATPs is by the way of \textbf{hammers}, which are proof assistant tools employing external ATPs to automatically find proofs of user given set of conjectures. Their main components are: lemmas (premises) selection which have high probability to be relevant to the set of conjectures, a translation between an ITP logic to a simpler ATP system, and then trials to prove the theorem by using combinations of existing theorems and search strategies \footnote{We should mention also a proof reconstruction module which translates a proof found by an ATP back to an ITP, so that it is accepted formally by an ITP as valid.}.

\medskip

\noindent Hammers are especially effective when one deals with ARLTs and have to juggle with hundreds of axioms and definitions. Their main goal is to make theorem proving more effective. For a general overview see \cite{hammer}; a hammer for COQ is described in \cite{hammer2} and for Mizar in \cite{ku}.   

\medskip

\noindent In the recent work \cite{deep} authors develop a deep learning hammer for Mizar and E prover. This is the very first use of deep learning techniques in theorem proving and already shows good results. We infer that similar techniques can be used in constructing a (semi)-automated dictionary. One of the keys to use machine learning in ITPs is constructing a dependency graph of definitions/theorems (cf. \cite{hk},\cite{hk2}). Thanks to its structure the Stacks Project has an easy to construct dependancy graph of its definitions, lemmas and theorems, which can be easily searched and hopefully also easily translated into COQ/Mizar, as we argued in the previous section. The easily accessible structure of the Stacks Project allows us to hope that one can recreate large parts of its content building on well-chosen premises selected at the beginning. This poses a natural problem in the spirit of recreating contents of COQ or Mizar Libraries (solved to some extent in \cite{deep}). The next step is to go beyond the Stacks Project and prove entirely new theorems, not previously proved by humans.

\section{Summary}

\noindent In this text we have outlined a program for possible formalization of large parts of algebraic geometry. Our plan can be summarized in the following steps:

\medskip

\noindent (1) Construct an (semi-)automated dictionary between (La)TeX and Interactive Theorem Provers, by exploiting existing syntactic parsers. 

\medskip

\noindent (2) Formalize The Stacks Project using this dictionary (and verify its correctness along the way).

\medskip

\noindent (3) Use automatic theorem provers with The Stacks Project as the input to prove new theorems in algebraic geometry and/or fill in gaps in \cite{stacks}. 

\medskip

\noindent We will report the progress on the implementation of this program in subsequent papers.

\end{document}